\documentclass{article}
\usepackage[T1]{fontenc}
\usepackage{spconf,amsmath,amssymb,graphicx,booktabs,xcolor,url,flushend}
\usepackage{tikz}
\usepackage[hidelinks]{hyperref}

\newlength{\resultwidth}
\AtBeginDocument{\settowidth{\resultwidth}{\small$100.00\pm10.00$}}

\newlength{\deltawidth}
\AtBeginDocument{\settowidth{\deltawidth}{\small$+10.00\,[-10.00,+10.00]$}}

\newcommand{\CpscFoneMean}{80.46}

\newcommand{\PtbxlFoneMean}{71.68}

\newcommand{\ChapmanFoneMean}{96.86}

\newcommand{\CpscTrainableM}{13.40}

\newcommand{\ECGJevName}{ECGJev}
\title{\ECGJevName: Question-Conditioned Option Scoring for 12-Lead ECG Interpretation}
\name{\begin{tabular}[b]{@{}c@{}}Zhuoyuan Li$^{1,\dagger}$, Xidong Wu$^{2,3,\dagger}$, Hanyu Cui$^{2,3}$, Jiansong Gao$^{2,3}$,\\
Rongrong Fu$^{2,3}$, Mingke Yan$^{4}$, Haotian Tang$^{2,3,*}$\end{tabular}}
\address{$^{1}$The Chinese University of Hong Kong, Hong Kong SAR; $^{2}$University of Dundee, Scotland, UK\\
$^{3}$Northeastern University, Shenyang, China; $^{4}$Peking University, Beijing, China\\
$^{\dagger}$Equal contribution; $^{*}$Corresponding author}
\begin{document}
\maketitle
\begin{abstract}
Clinical interpretation of a 12-lead electrocardiogram (ECG) often requires answering targeted diagnostic questions with clear, quantifiable decisions. We introduce ECGJev, the first ECG framework built on a decision decoder, a pretrained language model trained to return a probability for every listed option. A question-conditioned bridge compresses frozen waveform and spectrogram features into 16 ECG tokens, and a two-stage procedure adapts the bridge and low-rank adapters on the decoder's native option readout using diagnostic labels alone, scoring all options for a question in one forward pass. Evaluated across four ECG benchmarks, ECGJev achieves the highest macro-F1 on CPSC2018 (\CpscFoneMean\%), PTB-XL (\PtbxlFoneMean\%) and Chapman (\ChapmanFoneMean\%), and generalizes competitively to the small-sample PTB cohort. Concept probes and perturbation analyses further show that its predictions rely directly on the ECG input, remain stable under option-order and wording variations, and transfer zero-shot to unseen waveform concepts. Our code is available at \url{https://github.com/zhuoyuanli-lab/ECGJev}.
\end{abstract}

\begin{figure*}[t]
\centering
\includegraphics[width=\textwidth]{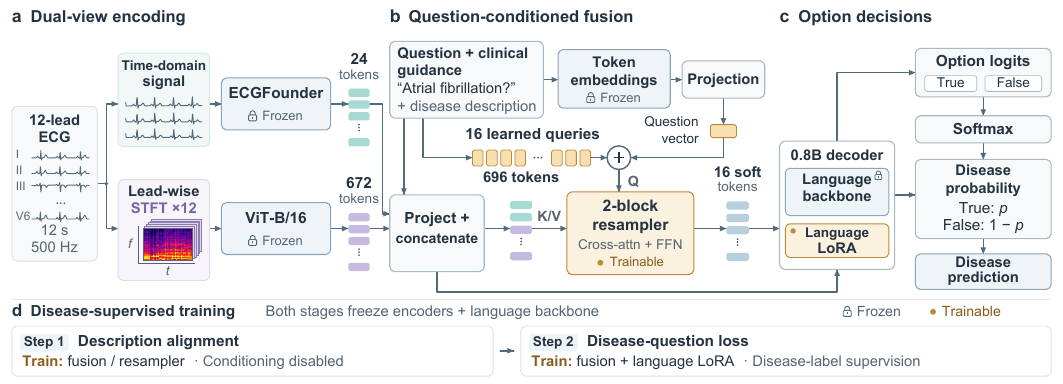}
\caption{ECGJev connects two views of the same recording to a diagnostic question: (a) frozen waveform and spectrogram-image encoders; (b) question-conditioned compression into 16 ECG tokens; (c) direct option scoring; and (d) description alignment followed by the disease-question loss. The clinical guidance is a textual description of the queried disease. Token counts illustrate CPSC2018.}
\label{fig:architecture}
\end{figure*}

\section{Introduction}
Interpreting a 12-lead electrocardiogram (ECG) relates beat timing, conduction intervals and wave morphology~\cite{beatssl2026}, motivating multiscale gating, lightweight and noise-routed models~\cite{wu2025,tang2026,wu2026mscgn,miao2026mdf} and foundation encoders~\cite{ribeiro2020,li2025}.

A spectrogram complements the waveform with local frequency content laid out for an image encoder; TSRNet~\cite{bui2024} combined both views for anomaly detection, yet adapting them as the queried finding changes remains open.

Language offers a flexible way to specify that finding. While classification heads on foundation encoders~\cite{li2025,xecg2025} score fixed label sets in one pass, they lock the output space and cannot interpret textual criteria or unseen queries; ECG-language encoders~\cite{merl2024,ked2024} match global descriptions without question-specific token selection. Conversely, generative ECG-LLMs~\cite{gem2025,pulse2024,pham2025,anyecg2025} accept open questions, but detour structured judgments through free-text generation---requiring post-hoc parsing while exposing predictions to vocabulary-wide competition and prompt biases.

We instead formulate \emph{direct option scoring} via a pretrained decision model. StartLux-Decision~\cite{startlux2026} trains language models directly to output probability distributions over lettered options rather than free text, providing a typed decision primitive adaptable from diagnostic labels alone. Prior ECG models have not explored this paradigm. Passing hundreds of signal and spectral tokens to the decoder while isolating query-relevant evidence---e.g., RR regularity vs.\ conduction intervals---motivates conditioning compression on the query \emph{before} option scoring~\cite{instructblip2023}.

ECGJev (Fig.~\ref{fig:architecture}) makes three contributions. (1) To our knowledge, it is the first ECG model built on a decision decoder: compressed ECG tokens enter a pretrained decision model whose native option readout is optimized directly from diagnostic labels, aligning training with inference. By decoupling displayed option letters from clinical option meanings under randomized option-order adaptation, it suppresses lexical token priors and position bias for unbiased diagnostic judgment. (2) We design a lightweight question-conditioned multimodal bridge that fuses frozen waveform and spectrogram features into 16 ECG tokens with low-rank decoder adapters (\CpscTrainableM M trainable parameters on CPSC2018) and no free-text supervision. (3) Across four benchmarks, ECGJev outperforms both foundation-encoder heads and generative ECG-LLMs on CPSC2018, PTB-XL and Chapman, delivering decisive option probabilities, option-order invariance, and zero-shot transfer to waveform concepts.

\section{Method}

\subsection{Task and ECG representations}
Given a 12-lead recording $x$ and a diagnostic question $q_d$ about label $d$, ECGJev scores the options True and False. Each question is accompanied by \emph{clinical guidance}, a short textual description of disease $d$. Questions are evaluated separately for multilabel prediction. For mutually exclusive rhythm classes, the largest True--False score difference determines the prediction.

Recordings are resampled to 500~Hz (12~s for CPSC2018, 10~s otherwise), notch-filtered at 50~Hz, band-pass filtered at 0.67--40~Hz, baseline-corrected and $z$-score normalized following established preprocessing protocols~\cite{wu2025,tang2026}. The waveform branch takes the pre-pooling features of frozen ECGFounder~\cite{li2025}: 24 tokens of width 1,024 for CPSC2018. For the spectral branch, each lead's short-time Fourier transform (Hann window 256, hop 32, FFT size 512) gives $\log(1+|Z|)$ over 0--40~Hz, clipped at training-set 1st/99th percentiles. The image is resized to $224\times224$ and encoded by a frozen ViT-B/16~\cite{dosovitskiy2021}. Discarding the CLS token and pooling its $14\times14$ patch grid to four frequency rows yields $14\times4=56$ tokens per lead, i.e.\ $12\times56=672$ tokens of width 768.

\subsection{Question-conditioned compression}
We project both views to width $d_b=256$ and concatenate their $N$ tokens to form $H\in\mathbb{R}^{N\times d_b}$. Fixed sinusoidal temporal and frequency encodings and learned lead and view embeddings identify their origins. We average the frozen decoder embeddings of the question and its clinical guidance to obtain $e_q$. Options and ECG tokens are excluded from this average.

Let $U\in\mathbb{R}^{16\times d_b}$ denote learned queries and $P_q$ a learned projection to width $d_b$. The initial queries are
\begin{equation}
Q_0=U+\alpha 1_{16}P_q(e_q)^\top,
\end{equation}
where $1_{16}$ is a column vector of ones that broadcasts the question vector to all queries, and the learned scalar $\alpha$ starts at 0.1. Two blocks of cross-attention to $H$ and feed-forward updates produce 16 compressed tokens. Projection and layer normalization map them to the decoder width, yielding $Z_d\in\mathbb{R}^{16\times1024}$. The token budget is thus fixed while the compression can adapt to the question.

\begin{table*}[t]
\centering\footnotesize
\caption{Classification comparison across four ECG benchmarks (\%, mean $\pm$ SD over three runs). F1 denotes macro-F1; Acc.\ denotes label-wise binary accuracy on multilabel tasks and four-class accuracy on Chapman. \textbf{Bold}/\underline{underline}: best/second-best score per column.}
\label{tab:peers}
\vspace{2pt}
\setlength{\tabcolsep}{1.6pt}
\begin{tabular}{@{}llrrrrrrrr@{}}
\toprule
& & \multicolumn{2}{c}{CPSC2018} & \multicolumn{2}{c}{PTB-XL} & \multicolumn{2}{c}{Chapman} & \multicolumn{2}{c}{PTB}\\
Model & Setting & F1 $\uparrow$ & Acc. $\uparrow$ & F1 $\uparrow$ & Acc. $\uparrow$ & F1 $\uparrow$ & Acc. $\uparrow$ & F1 $\uparrow$ & Acc. $\uparrow$\\
\midrule
xECG~\cite{xecg2025} & head & $75.14\pm0.13$ & \underline{$95.57\pm0.07$} & $70.52\pm0.31$ & $86.65\pm0.37$ & \underline{$95.39\pm1.10$} & \underline{$96.84\pm0.63$} & $\mathbf{43.42\pm6.30}$ & $90.09\pm2.65$\\
MERL~\cite{merl2024} & head & $65.24\pm0.29$ & $92.15\pm0.35$ & $69.15\pm0.18$ & $85.51\pm0.21$ & $88.23\pm0.83$ & $91.43\pm0.38$ & $21.20\pm0.23$ & $59.78\pm4.07$\\
Q-Heart~\cite{pham2025} & adapter & \underline{$76.06\pm1.07$} & $95.50\pm0.18$ & \underline{$71.22\pm0.41$} & $\mathbf{88.78\pm0.13}$ & $91.18\pm0.27$ & $94.79\pm0.12$ & $14.82\pm0.80$ & $\mathbf{94.40\pm0.00}$\\
anyECG-chat~\cite{anyecg2025} & LoRA & $70.02\pm1.90$ & $95.25\pm0.08$ & $62.24\pm3.55$ & $87.28\pm0.25$ & $93.42\pm0.47$ & $96.10\pm0.36$ & $13.23\pm2.99$ & \underline{$94.22\pm0.41$}\\
\textbf{ECGJev} & ours & $\mathbf{80.46\pm0.67}$ & $\mathbf{96.68\pm0.16}$ & $\mathbf{71.68\pm0.38}$ & \underline{$87.52\pm0.41$} & $\mathbf{96.86\pm0.74}$ & $\mathbf{98.23\pm0.34}$ & \underline{$35.02\pm8.68$} & $83.69\pm4.67$\\
\bottomrule
\end{tabular}
\end{table*}

\begin{table}[t]
\centering\footnotesize
\caption{Ablation study on CPSC2018 (\%, mean $\pm$ SD). $\Delta$: performance gain of full ECGJev over each ablated variant (percentage points). \textbf{Bold}/\underline{underline}: best/second-best scores.}
\label{tab:ablations}
\vspace{2pt}
\setlength{\tabcolsep}{2pt}
\begin{tabular}{@{}lrrrr@{}}
\toprule
& \multicolumn{2}{c}{Macro-F1} & \multicolumn{2}{c}{Accuracy}\\
Variant & Score & $\Delta$ & Score & $\Delta$\\
\midrule
\textbf{ECGJev} & $\mathbf{80.46\pm0.67}$ & \textemdash & $\mathbf{96.68\pm0.16}$ & \textemdash\\
Spectrogram only & $64.57\pm0.73$ & $+15.89$ & $92.45\pm0.52$ & $+4.23$\\
w/o conditioning & $79.75\pm1.04$ & $+0.71$ & $96.52\pm0.12$ & $+0.16$\\
Name-only & \underline{$79.98\pm1.02$} & $+0.48$ & $96.55\pm0.16$ & $+0.13$\\
Length control & $79.60\pm0.83$ & $+0.86$ & \underline{$96.59\pm0.08$} & $+0.09$\\
w/o alignment & $79.04\pm1.58$ & $+1.42$ & $96.51\pm0.39$ & $+0.17$\\
\bottomrule
\end{tabular}
\end{table}

\subsection{Option scoring and adaptation}
\emph{Decision decoder.} We adopt the 0.8B StartLux-Decision checkpoint~\cite{startlux2026}, a 24-layer Qwen3.5-based architecture mixing linear and full attention at width 1,024. Pretrained with soft cross-entropy over option-letter probabilities across 2.7M general-domain text decisions (classification, question answering, safety and tool use), it provides a dedicated option-scoring readout prior to physiological-signal adaptation. The tokens $Z_d$ occupy the decoder's continuous-input positions alongside the question, clinical guidance and lettered options. Let $h_{\rm last}$ be its hidden state at the last non-padding prompt position. If $\pi(c)$ maps option meaning $c$ to its displayed letter and $w_{\pi(c)}$ is the corresponding frozen output-head vector, then
\begin{equation}
z_c=w_{\pi(c)}^\top h_{\rm last},\quad
p_d=\tfrac{e^{z_{\rm T}}}{e^{z_{\rm T}}+e^{z_{\rm F}}}=\sigma(\Delta_d),\ \
\Delta_d=z_{\rm T}-z_{\rm F}.
\end{equation}
Here $p_d$, the softmax probability of True over the two options, equals the sigmoid $\sigma$ of the margin $\Delta_d$; letters are mapped back to meanings. Each question requires one causal forward pass and no answer-token generation.

\emph{Unbiased option scoring via symbol decoupling.} Generative models such as Q-Heart and anyECG-chat approximate decisions via ``yes''/``no'' token likelihoods, where both tokens compete across the full vocabulary under a text-generation objective, inheriting lexical frequency priors and prompt-order sensitivity. Conversely, fixed classification heads cannot read clinical criteria or score unseen concepts. ECGJev overcomes both limitations: Stage~2 optimizes the exact native decision margin used at inference while mapping arbitrary option letters back to clinical meanings via $\pi(c)$ under random letter-order swaps. Decoupling surface symbols from diagnostic semantics suppresses lexical and position biases while unifying multilabel screening, mutually exclusive rhythm selection and zero-shot concept probing under one typed readout (Sec.~\ref{sec:behavior}).

\emph{Stage 1 (description alignment)} trains the bridge without question conditioning. For record $i$ and disease $d$, the logit $s_{id}=\cos(\bar z_i,g_d)/\tau+b_d$ compares the mean compressed token $\bar z_i$ with the frozen embedding $g_d$ of the guidance for $d$; $\tau=0.07$, and the learned bias $b_d$ is initialized from training prevalence. Annotated-present labels are positives, annotated-absent labels negatives, and unknown labels are masked in a binary cross-entropy loss (10 epochs, batch 32, learning rate $3\times10^{-4}$). \emph{Stage 2 (disease-question loss)} applies the same masked binary cross-entropy to $\Delta_d$, equivalent to cross-entropy over the two options. It updates the bridge and language-layer low-rank adapters (LoRA, rank 16)~\cite{hu2021}, 13.40M parameters on CPSC2018; the encoders, decoder base and output head remain frozen. AdamW uses bridge/LoRA learning rates $10^{-4}$/$5\times10^{-5}$ and batch 8 for at most 8 epochs, with early stopping and checkpoint selection by validation average precision (AP). Training covers all target questions per record, each drawn from two of three fixed wordings per disease with the option order swapped at random; evaluation uses one fixed wording.

\section{Experiments and results}
\subsection{Datasets and evaluation protocol}
\emph{Data.} Following established preprocessing and stratified partitioning protocols~\cite{wu2025,tang2026}, each dataset is split 70\% / 15\% / 15\% (train / validation / test): CPSC2018~\cite{liu2018} with nine diagnostic labels (4,814 / 1,032 / 1,031); PTB-XL~\cite{wagner2020,strodthoff2023,goldberger2000} with five diagnostic superclasses (15,258 / 3,270 / 3,270); Chapman~\cite{zheng2020} with four mutually exclusive rhythm classes---atrial fibrillation, supraventricular tachycardia, sinus bradycardia and sinus rhythm (6,082 / 1,296 / 1,299); and PTB~\cite{ptb2004,goldberger2000} (350 / 75 / 75) across ten shared diagnostic codes.

\emph{Metrics.} Macro-F1 weights classes equally, and multilabel accuracy measures label-wise binary accuracy. Multilabel outputs are binarized at validation-optimal F1 thresholds (four-class argmax on Chapman); values are mean $\pm$ SD over three runs.

\emph{Comparators.} All baselines are adapted per training split with validation-selected checkpoints. Foundation encoders xECG~\cite{xecg2025} and MERL~\cite{merl2024} train classification heads on frozen pooled features. Q-Heart~\cite{pham2025} fine-tunes task adapters (no-retrieval setting) and scores the summed ``yes''--``no'' log-likelihood margin; anyECG-chat~\cite{anyecg2025} fine-tunes its multimodal bridge and LoRA adapters (23.3M parameters, frozen 8B backbone) on the first-token ``yes''--``no'' log-probability margin.

\subsection{Comparison across four datasets}
\label{sec:cmp}
\begin{figure}[!t]
\raggedright
\input{generated/question_answering_comparison}
\caption{Two representative CPSC2018 validation ECGs (lead II waveform and 0--40~Hz spectrogram; gray denotes padding). ECGJev uses all 12 leads and outputs explicit two-option probabilities; Q-Heart and free-text anyECG-chat responses are verbatim.}
\label{fig:qa}
\end{figure}
Table~\ref{tab:peers} shows that ECGJev attains the highest macro-F1 on CPSC2018 (80.46\%), PTB-XL (71.68\%) and Chapman (96.86\%) across multilabel and exclusive rhythm tasks, beating the best baseline by 4.40 F1 points on CPSC2018 (96.68\% accuracy) and 1.47 on Chapman (98.23\% accuracy), and 8B anyECG-chat by 9.44 points on PTB-XL (87.52\% accuracy). On small-sample PTB, where imbalance inflates negative accuracy, generative baselines collapse to negative priors (13.23\%--14.82\% F1), whereas ECGJev's decision prior reaches 35.02\%---more than doubling both LLMs.

\subsection{Component analysis}
Table~\ref{tab:ablations} ablates key components on CPSC2018 (16 ECG tokens). Full ECGJev attains the highest macro-F1 (80.46\%) and accuracy (96.68\%) with the lowest SD (0.67\% vs.\ 0.73\%--1.58\%). Using spectrograms alone lowers F1 by 15.89 points, showing that waveforms carry the primary morphology signal while spectrograms add complementary time--frequency cues. Omitting Stage~1 alignment ($\Delta=+1.42$, $\text{SD}=1.58\%$), removing conditioning ($Q_0=U$, $\Delta=+0.71$), or replacing clinical guidance with disease names ($\Delta=+0.48$) or length-matched text ($\Delta=+0.86$) consistently lowers F1 and increases variance.

\subsection{Question-specific behavior and interpretation}
\label{sec:behavior}
In Fig.~\ref{fig:qa}, ECGJev and Q-Heart match all four annotations, while ECGJev directly grades each option (True 0.891 and 0.982 on positives vs.\ $\le0.002$ otherwise); free-text anyECG-chat hedges on record~2 without a clear decision.

\begin{table}[t]
\centering\footnotesize
\caption{Zero-shot waveform concept probing on PTB-XL~\cite{wagner2020,strodthoff2023}. AUROC is averaged over three runs; $n_+$ denotes positive test instances per annotated concept~\cite{lee2026}.}
\label{tab:concepts}
\vspace{2pt}
\begin{tabular}{@{}lrr@{}}
\toprule
Queried concept & $n_+$ & AUROC $\uparrow$\\
\midrule
Prolonged PR & 155 & 0.755\\
Irregular RR & 227 & 0.743\\
ST depression & 151 & 0.895\\
ST elevation & 4 & 0.605\\
Inverted T waves & 44 & 0.883\\
\bottomrule
\end{tabular}
\end{table}
Following prior ECG concept-probing protocols~\cite{strodthoff2023,lee2026}, Table~\ref{tab:concepts} evaluates zero-shot transfer to five unseen waveform concepts on PTB-XL~\cite{wagner2020}. Trained on superclass labels alone, ECGJev generalizes well to repolarization and conduction queries (0.895 AUROC on ST depression, 0.883 on inverted T waves, and 0.755/0.743 on prolonged PR and irregular RR), with lower sensitivity only on rare ST elevation ($n_+=4$).

Perturbation tests confirm symbol--meaning decoupling and signal grounding: reversing True/False letters shifts mean AP by at most 0.30 points across PTB-XL, Chapman and PTB (near-zero position bias), whereas zeroing ECG tokens drops PTB-XL AP from 77.20\% to 25.42\%, verifying reliance on ECG input rather than prompt priors.

\section{Discussion and conclusion}
By coupling frozen dual-view ECG encoders with a pretrained decision decoder, ECGJev turns clinical queries into composable, unbiased decision primitives. Unlike closed-set linear heads or multi-billion-parameter chat MLLMs prone to lexical priors, position bias and post-hoc parsing, our 0.8B framework adapts from routine labels alone (\CpscTrainableM M trainable parameters) to return explicit option probabilities in one forward pass---allowing downstream workflows to tune operating thresholds or escalate ambiguous cases without re-prompting, while transferring zero-shot to waveform concepts. Future work will refine clinical query formulations and strengthen the model's judgment of detailed ECG feature descriptions for more reliable diagnostic decision-making.

In conclusion, ECGJev connects 12-lead ECG waveforms and spectrograms to a decision decoder via 16 question-conditioned tokens, symbol-decoupled option scoring and two-stage label adaptation, achieving superior classification across CPSC2018, PTB-XL and Chapman with stable, directly actionable diagnostic probabilities.

\clearpage
\section{Compliance with ethical standards}
This study used only public, de-identified ECG datasets~\cite{liu2018,wagner2020,zheng2020,ptb2004}; no new data were collected. The authors declare that they have no known competing financial interests or personal relationships that could have appeared to influence the work reported in this paper.
\begingroup
\ninept 
\interlinepenalty=10000
\bibliographystyle{IEEEbib}
\bibliography{references}
\endgroup
\end{document}